\documentclass[final]{cvpr}

\usepackage{times}
\usepackage{epsfig}
\usepackage{graphicx}
\usepackage{amsmath}
\usepackage{amssymb}

\usepackage{multirow}
\usepackage{booktabs}
\usepackage{makecell}
\usepackage{setspace}
\usepackage{bm}

\usepackage[pagebackref=true,breaklinks=true,colorlinks,bookmarks=false]{hyperref}
\usepackage{nopageno}

\newcommand{\ra}{\rightarrow} % Use \ra to replace \rightarrow

\def\cvprPaperID{3147} % *** Enter the CVPR Paper ID here
\def\confYear{CVPR 2021}

\begin{document}

%%%%%%%%% TITLE
\title{EffiScene: Efficient Per-Pixel Rigidity Inference for Unsupervised Joint Learning of Optical Flow, Depth, Camera Pose and Motion Segmentation}

\author{
Yang Jiao $^{*1,2,3}$, Trac D. Tran $^{2}$, Guangming Shi $^{1,3}$\\
$^{1}$ Xidian University, $^{2}$ Johns Hopkins University, $^{3}$ Xidian Guangzhou Institute of Technology\\
\tt\small \{yangjiao,yjiao8\}@\{stu.xidian.edu.cn,jhu.edu\}, trac@jhu.edu, gmshi@xidian.edu.cn
}
\maketitle

\thispagestyle{empty}

%%%%%%%%% ABSTRACT
\begin{abstract}

    This paper addresses the challenging unsupervised scene flow estimation problem by jointly learning four low-level vision sub-tasks: optical flow $\textbf{F}$, stereo-depth $\textbf{D}$, camera pose $\textbf{P}$ and motion segmentation $\textbf{S}$. Our key insight is that the rigidity of the scene shares the same inherent geometrical structure with object movements and scene depth. Hence, rigidity from $\textbf{S}$ can be inferred by jointly coupling $\textbf{F}$, $\textbf{D}$ and $\textbf{P}$ to achieve more robust estimation. To this end, we propose a novel scene flow framework named EffiScene with efficient joint rigidity learning, going beyond the existing pipeline with independent auxiliary structures. In EffiScene, we first estimate optical flow and depth at the coarse level and then compute camera pose by Perspective-$n$-Points method. To jointly learn local rigidity, we design a novel Rigidity From Motion (RfM) layer with three principal components: \emph{}{(i)} correlation extraction; \emph{}{(ii)} boundary learning; and \emph{}{(iii)} outlier exclusion. Final outputs are fused based on the rigid map $M_R$ from RfM at finer levels. To efficiently train EffiScene, two new losses $\mathcal{L}_{bnd}$ and $\mathcal{L}_{unc}$ are designed to prevent trivial solutions and to regularize the flow boundary discontinuity. Extensive experiments on scene flow benchmark KITTI show that our method is effective and significantly improves the state-of-the-art approaches for all sub-tasks, i.e. optical flow ($5.19 \ra 4.20$), depth estimation ($3.78 \ra 3.46$), visual odometry ($0.012 \ra 0.011$) and motion segmentation ($0.57 \ra 0.62$).

\end{abstract}

%%%%%%%%% BODY TEXT
%%%%%%%%% ==> Introduction
\section{Introduction}
\label{sec:1}
Scene flow \cite{SceneFlow99, SceneFlow05} describes the 3D motion of a dynamic scene by 2D optical flow and scene depth, providing essential geometrical clues for numerous practical applications such as self-driving \cite{Menze} and robotics navigation \cite{Robot, Robot2}. However, acquiring dense ground truth for both sub-tasks in real applications are usually expensive or impractical. To overcome this, learning scene flow in an unsupervised way has attracted much attention in recent years, by minimizing the photometric differences between the original-synthesized pixel pairs.

Optimizing pixel-wise photometric error for low-level scene flow task without supervision is not a trivial task. One of the most critical reason is that the pixel correspondence between consecutive frames is highly ambiguous, especially in unstructured or texture-less regions. For example, one pixel from a mountain or a highway surface in frame $t$ can be projected to various surrounding pixels in frame $t+1$ with very low photometric error, often leading to the failure of local scene flow estimation. Unfortunately, this issue always happens in outdoor scenarios due to missing small details due to motion blur. Therefore, additional constraints are strongly needed to eliminate the ambiguities for successful unsupervised scene flow estimation.

\begin{figure}[htbp]
\begin{center}
    \includegraphics[width=1.0\linewidth]{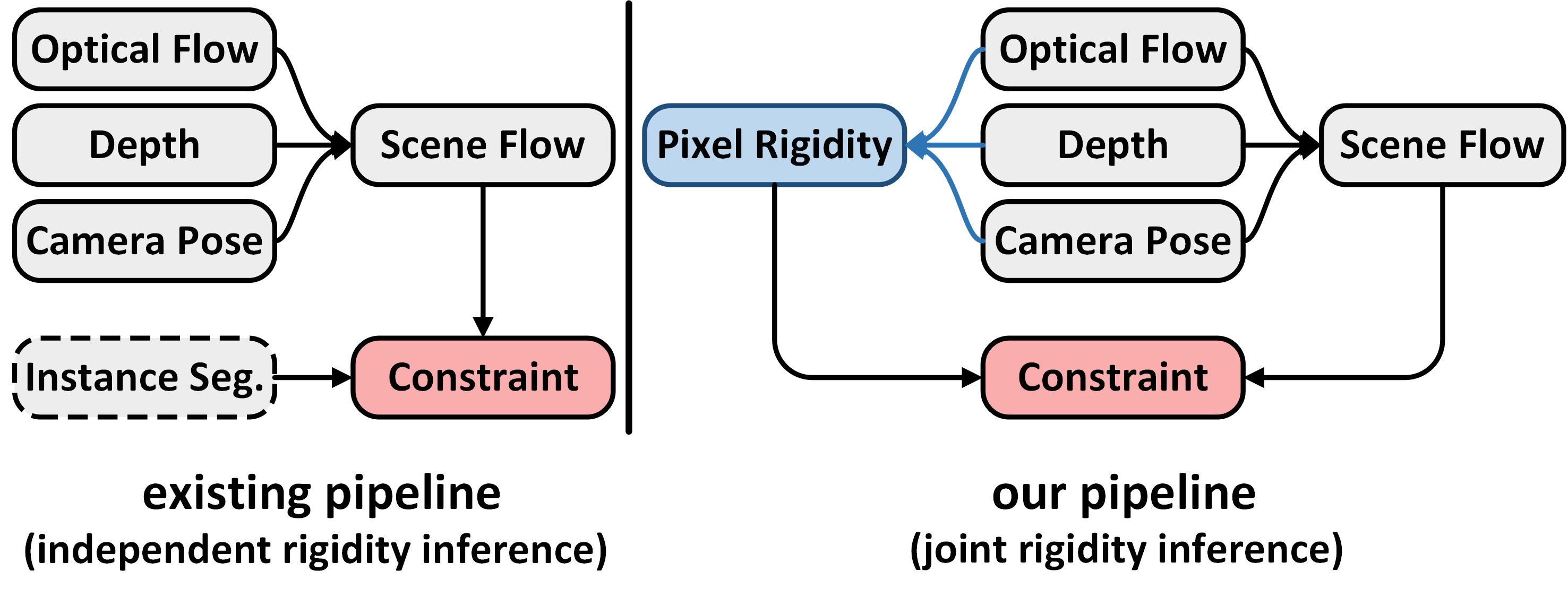}
\end{center}
    \caption{Main idea of our method. Different from independently estimating rigid pixels from the auxiliary instance segmentation in existing pipeline, we jointly learn per-pixel rigidity from optical flow, depth and camera pose for more accurate rigid constraint.}
\label{fig:DiffWithOther}
\end{figure}

\textbf{Problems.} In recent approaches, rigid constraint is widely employed to separate the scene into static (rigid) and moving (non-rigid) areas. It also restricts the ego-motion of the rigid pixels which obey the rigid scene assumption \cite{RigidAssumption}. To achieve this, current methods \cite{ CC, DeepRigid, EPC, SENSE, Hu2020, SIGNet} follow a popular scene flow pipeline as shown in Fig. \ref{fig:DiffWithOther} (left), where the auxiliary instance segmentation network is designed to predict the rigid pixels that will be constrained by local rigidity. Though impressive scene flow results can be achieved, the performance of the segmentation is often poor, indicating an inaccurate estimation of rigid pixels (static area), and it could, in turn, harm the rigid constraint. One reason is that the independent rigidity inference in existing pipeline limits the learning of pixel rigidity. More specifically, the segmentation task in existing pipeline is jointly optimized with scene flow sub-tasks in back-propagation, but it is independently launched in forward inference. This independent structure makes inference inefficient, resulting in networks that can only learn pixel-wise rigidity from raw RGB images, but difficult to extract extra geometrical information from flow and depth. Besides, optimizing both deep segmentation network and scene flow multi-networks in current pipeline without ground truth might be very difficult, requiring sophisticated training strategies such as \cite{CC}.

\textbf{Motivation \& Idea.} Inspired from recent works, our key insight is that the rigidity of the scene shares the same inherent geometrical structure with optical flow and depth, hence they are highly correlated and can be mutually beneficial. Based on this observation, instead of designing the auxiliary segmentation structure, we jointly consider optical flow, depth and camera pose for rigidity learning as illustrated in Fig. \ref{fig:DiffWithOther} (right), and propose a novel framework called EffiScene. With the new pipeline, we can go beyond the existing methods by providing: \emph{(i)} more effective rigid constraint via jointly considering scene flow sub-tasks for learning accurate rigid pixels; and \emph{(ii)} more efficient scene flow framework optimization via eliminating the very deep instance segmentation network.

\textbf{Approach.} EffiScene aims to solve the following four unsupervised sub-tasks: \emph{(i)} optical flow $\textbf{F}$ estimation; \emph{(ii)} stereo-depth $\textbf{D}$ prediction; \emph{(iii)} camera pose $\textbf{P}$ for visual odometry; and \emph{(iv)} motion segmentation $\textbf{S}$. We first estimate the optical flow $F^o$ and depth $D$ at the coarse level, then compute the relative camera pose $P$ from time $t$ to $t+1$ by minimizing the reprojection error between the observed coordinates (from $F^o$) and the projected 3D points (from $D$) via a Perspective-$n$-Points (P$n$P) solver. Next, we propose a novel Rigidity From Motion (RfM) layer to estimate pixel rigidity by explicitly modeling the correlation between optical flow $F^o$ and rigid flow $F^r$. Our RfM includes three main steps: \emph{(i)} correlation extraction; \emph{(ii)} boundary learning; and \emph{(iii)} outlier exclusion. The rigid map $M_R$ from RfM can be interpreted as motion segmentation. Finally, flows from $F^o$ and $F^r$ are fused to form the final flow, guided by the rigid map $M_R$ at the fine level. In training, two new losses -- $\mathcal{L}_{bnd}$ and $\mathcal{L}_{unc}$ -- are designed to optimize RfM and regularize the flow boundary discontinuity, respectively. Different from existing methods \cite{UnOS, UnRigidFlow}, there are no sensitive thresholds needed to be set manually in EffiScene.

\textbf{Contributions} are summarized as follows.
\begin{itemize}

\item We introduce a new structure for unsupervised scene flow estimation, and demonstrate that per-pixel rigidity can be efficiently predicted by jointly learning optical flow, depth and camera pose.

\item We design a novel Rigidity from Motion (RfM) layer to recognize rigid regions via explicitly modeling motion correlations. To the best of our knowledge, this is the first deep model for joint rigidity learning.

\item We optimize scene flow training by two new losses: $\mathcal{L}_{bnd}$ prevents the trivial solution of RfM whereas $\mathcal{L}_{unc}$ regularizes the optical flow discontinuity in uncovered boundary.

\end{itemize}

Extensive experiments on KITTI benchmarks \cite{KITTI_raw, KITTI_2012, KITTI_2015} show that our method outperforms existing state-of-the-art (SOTA) approaches for all four sub-tasks with highly efficient rigidity inference (RfM with size 0.0032Mb vs. 5.22Mb \cite{CC}), i.e. optical flow ($5.19 \ra 4.20$) by a significant $19\%$ improvement, depth estimation ($3.78 \ra 3.46$), visual odometry ($0.012 \ra 0.011$), and motion segmentation ($0.57 \ra 0.62$).

%%%%%%%%% ==> Related Works
\section{Related Work}
\label{sec:2}
We first offer a brief review of optical flow and depth estimation, which are jointly learned for efficient rigidity inference in our method. Then, scene flow is discussed.

\textbf{Optical flow.} Deep convolutional neural networks (CNN) are widely used in supervised optical flow methods. FlowNet \cite{FlowNet} is the first work using an end-to-end CNN architecture. FlowNet2 \cite{FlowNet2} improves the results by stacking more layers but can be computationally expensive. Then, simpler deep models such as SpyNet \cite{SPyNet}, LiteFlowNet \cite{LiteFlowNet} and PWC-Net \cite{PWC-Net} are designed in spatial or feature pyramid fashion. Very recently, recurrent units are designed for decoding all-pair cost volumes in RAFT \cite{RAFT} and it achieves state-of-the-art result. These works \cite{UnFlow, DDFlow, SelFlow, Janai, ARFlow} provide effective backbones for unsupervised methods in which flow is learned by optimizing photometric loss from synthesis views \cite{Back2Basics}. To address occlusions and large displacements, novel losses and training strategies are designed, such as bidirectional census loss in UnFlow \cite{UnFlow}, data distillation in DDFlow \cite{DDFlow} and SelFlow \cite{SelFlow}, and extra forward pass in ARFlow \cite{ARFlow}.

Different from these works, we construct the final optical flow by fusing the  motion from moving and static area guided by the learned rigid map $M_R$.

\textbf{Depth estimation.} Comparing with monocular estimation, learning depth from stereo images in the absence of the ground truth provides higher quality results. The self-supervision signal comes from the left-right synthesis view. In stereo works, Garg \etal \cite{Garg} firstly adopt an auto-encoder to predict continuous values of disparity. Godard \etal \cite{MonoDepth} introduce a left-right consistency term for geometry constraint, then improves it by associated design choices \cite{MonoDepth2}. Temporal information is also considered in \cite{UnDeepVO, DeepVO}.
In monocular based methods, since single-view depth is usually insufficient for self supervision, extra information is borrowed from consecutive frames \cite{sfMLearner, SfM-Net, Vid2Depth}. 

\textbf{Unsupervised scene flow estimation.} Traditional scene flow techniques have achieved impressive results, usually at high computational cost, such as super-pixel scene decomposition \cite{Menze} and Plane+Parallax framework \cite{MR-Flow}. Even the fast version \cite{FSF} still runs in 2-3 seconds per frame. In deep model, GeoNet \cite{GeoNet} implicitly represents moving pixels by refining the residual non-rigid flow via ResFlowNet, and DF-Net \cite{DF-Net} imposes a cross-task consistency loss for rigid area. Recently, per-pixel rigid constraint is adopted by incorporating deep segmentation network into scene flow. For instance, Ma \etal \cite{DeepRigid} adopt off-the-shelf Mask R-CNN \cite{MaskRCNN} for rigid instance segmentation, and Yang \etal \cite{EPC} predict moving masks via MotionNet followed by a holistic 3D motion parser (HMP). To simplify the multi-task training, Ranjan \etal \cite{CC} present Collaborative Competitive (CC) to facilitate the network coordination. Wang \etal \cite{UnOS} yield static pixels based on the flow residual, and Liu \etal \cite{UnRigidFlow} extend it via local rigidity. However, both are sensitive to thresholds, which may lead to the inaccurate motion area.

These methods achieve very impressive results, but the ego-motion is constrained in low efficiency due to the independent inference process. In our framework, we efficiently learn per-pixel rigidity by jointly coupling optical flow, depth and camera pose, leading to considerable improvements of each sub-task.

%%%%%%%%% ==> Methods
\section{EffiScene Method}
\label{sec:3}
We first introduce the preliminary geometrical rigid consistency in Sec. \ref{sec:3.1}, and then describe the design of RfM in Sec. \ref{sec:3.2}. To present the new pipeline, we first construct the overall algorithm in Sec. \ref{sec:3.3}, and discuss the design of the new loss as well as regularization functions in Sec. \ref{sec:3.4}.

% === Section 3.1 === %
\subsection{Geometrical Rigid Consistency}
\label{sec:3.1}
Given two consecutive frames $I_t$ and $I_{t+1}$, pixel movements can be divided into two categories: \emph{(i)} local motion from moving objects denoted by optical flow $F^{o}_{t\ra{t+1}}$; and \emph{(ii)} global (or ego) motion from backgrounds described by rigid flow $F^{r}_{t\ra{t+1}}$. Compared with the optical flow $F^{o}_{t\ra{t+1}}$, rigid flow $F^{r}_{t\ra{t+1}}$ strictly follows the rigid geometrical consistency, hence it has lower corresponding ambiguity.

Rigid flow field $F^{r}_{t\ra{t+1}}$ can be easily determined in 2D cases such as FlyingChairs \cite{FlowNet}, where the depth $D$ is treated as a constant, by applying a 4-DoF plane affine transformation $P$. However, in realistic 3D scenes like KITTI suite \cite{KITTI_raw}, the geometrical consistency of global motion can be only guaranteed by re-projecting 2D points $\mathbf{x}$ back to 3D world coordinates $\mathbf{X}$ with perspective transformation:
\begin{equation}
    [\mathbf{x}; 1] = \frac{1}{d} KPM [\mathbf{X}; 1],
\label{eq:CameraModel}
\end{equation}
where $K \in \mathbb{R}^{3\times3}$ is the camera intrinsic parameter, $P \in \mathbb{R}^{3\times4}$ stands for the relative camera pose and $M \in \mathbb{R}^{4\times4}$ indicates the object movement in world coordinate system. Also, $d$ is the normalization coefficient for $3D \rightarrow 2D$ projection, and it indicates the per-pixel depth for $\mathbf{x}$. Based on this, the movement of the static region (with $M=\boldsymbol{I}$) is computed by the differences between $\mathbf{x}_{t}$ and $\mathbf{x}_{t+1}$, and can be formulated via the rigid flow $F^r_{t\ra{t+1}}$ as shown below
\begin{equation}
\setlength{\abovedisplayskip}{3pt}
\setlength{\belowdisplayskip}{3pt}
\begin{aligned}
    F^r_{t\rightarrow{t+1}} = \frac{1}{d_{t+1}} KP (d_{t} K^{-1} \mathbf{x}_{t}) - \mathbf{x}_{t}.
\label{eq:RigidFlow}
\end{aligned}
\end{equation}
Here, the depth $d_t$ of the source image $I_t$ and the camera pose $P$ are the only two unknowns to be estimated in the computation of $F^r_{t\ra{t+1}}$. With this geometrical rigid consistency, global motion from (\ref{eq:RigidFlow}) are jointly considered with local motion in RfM for efficient rigid area recognition.

% === Section 3.2 === %
\subsection{Rigidity from Motion (RfM)}
\label{sec:3.2}
Instead of independently predicting pixel rigidity from raw images as in \cite{DeepRigid, EPC, CC}, we classify the static rigid area based on optical flow $F^o_{t\ra{t+1}}$ and rigid flow $F^r_{t\ra{t+1}}$. This is similar to \cite{UnOS, UnRigidFlow} where a simple threshold is used to binary divide static and moving regions. However, the essential difference in our method comes from the consideration that global motion in $F^r_{t\ra{t+1}}$ is strictly restricted by the geometrical rigid constraint whereas local motion from $F^o_{t\ra{t+1}}$ is free. The static area can be naturally determined by regarding the local motion as the outliers of the global motion. However, finding 3D motion outliers from 2D flow field is non-trivial since \emph{(i)} different 3D motion tensors can be represented by the same 2D flow vector; and \emph{(ii)} the learned flow filed itself might be inaccurate. Hence, we need to design the Rigidity from Motion (RfM) layer to adaptively learn the motion boundary by explicitly modeling the correlation of various flows.

% => Figure: Segmentation demo.
\begin{figure}[htbp]
\begin{center}
    \includegraphics[width=0.9\linewidth]{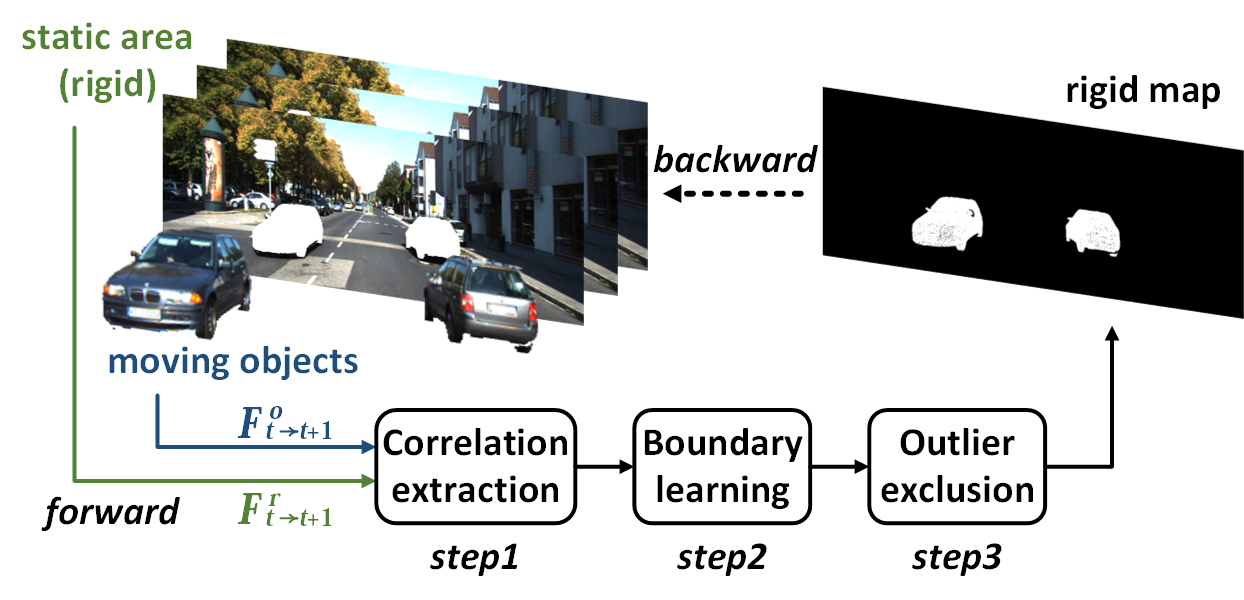}
\end{center}
   \caption{Illustration of RfM. (Flows are replaced by RGBs, and rigid map is shown in inverted colors for better visualization.)}
\label{fig:RfM}
\vspace{-4mm}
\end{figure}

\begin{figure}[htbp]
\begin{center}
\includegraphics[width=1.00\linewidth]{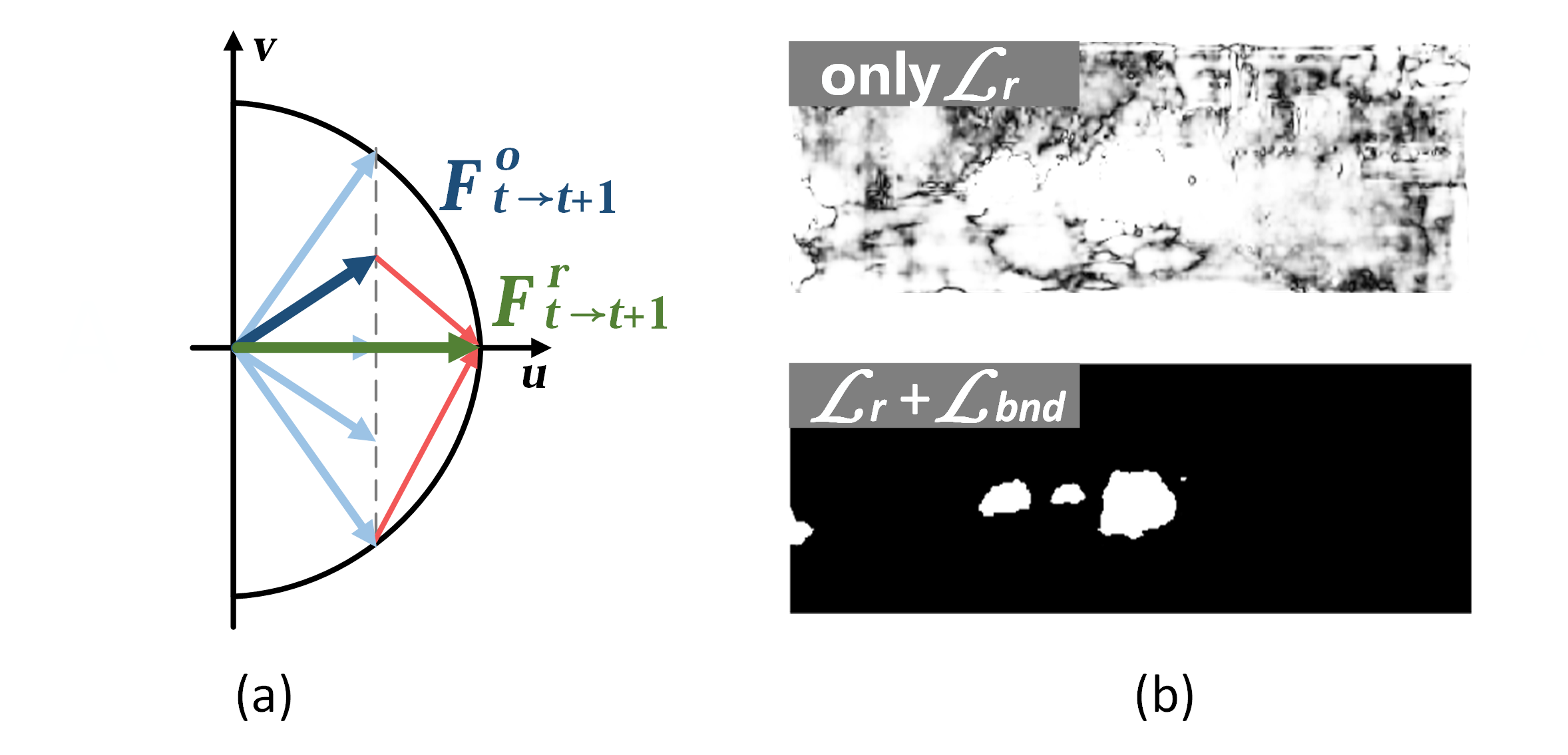}
\end{center}
   \caption{Illustration of (a) flow correlation and (b) effectiveness of boundary loss $\mathcal{L}_{bnd}$ to prevent trivial solutions.}
\label{fig:BndLoss}
\end{figure}

$\textbf{Forward}$. The RfM involves three steps as illustrated in Fig. \ref{fig:RfM}: \emph{(i)} correlation extraction; \emph{(ii)} boundary learning; and \emph{(iii)} outlier exclusion. In the first place, we construct a pixel wise correlation map $C_{F}$ as follows:
\begin{equation}
    C_{F} = \mathcal{N}(f_c(F^o_{t\rightarrow{t+1}}, F^r_{t\rightarrow{t+1}})),
\label{eq:RfM:step1}
\end{equation}
in which $f_c$ evaluates the per-pixel similarity between $F^o_{t\rightarrow{t+1}}$ and $F^r_{t\rightarrow{t+1}}$, while the operator $\mathcal{N}$ normalizes the correlation value to $[0, 1]$. Usually, inner product can be the first choice to evaluate the correlation between any two vectors. However, it could be insufficient to distinguish the flow in the case that a pixel moves very slowly (near zero) along one direction. For example, in Fig. \ref{fig:BndLoss} (a), the inner product between green motion $F^r_{t\ra{t+1}}$ and different blue motions $F^o_{t\ra{t+1}}$ will yield the same results regardless of the $v$-axis moving of $F^o_{t\ra{t+1}}$. To avoid this issue, we rely on the intuitive but more effective $l_2$-norm for motion residual (red arrows) to describe the motion similarity by $f_c = {||F^o_{t\rightarrow{t+1}}- F^r_{t\rightarrow{t+1}}||}_2$.
$C_{F}$ evaluates the similarity between the two flows $F^o_{t\rightarrow{t+1}}$ and $F^r_{t\rightarrow{t+1}}$, which are more similar (with $C_{F}=0$) at rigid region, while dissimilar (with $C_{F}=1$) at non-rigid region. According to the Central Limit Theorem, the distribution of $C_{F}$ in rigid region can be seen as a Gaussian with mean value 0, while non-rigid region fits another Gaussian with mean value 1. Based on this, secondly, we compute the overall histogram $h_F$ of $C_F$, and naturally separate rigid and non-rigid pixels from the histogram by a Gaussian Mixture Model (GMM). To enforce differentiability, we approximate the GMM by designing a fully connected network $g(h_F|\theta)$ with learnable parameter $\theta$. $g(h_F|\theta)$ automatically regresses the optimal rigid boundary by learning from the input $h_F$. And lastly, we construct the rigid map $M_R$ in (\ref{eq:alg:step3}) to exclude local motion outliers from the global motion.
\begin{equation}
    M_R = 1 - 1 / (1 + \alpha \cdot (C_F - g(h_F|\theta))).
\label{eq:alg:step3}
\end{equation}
In this equation, $\alpha$ controls the balance between "hard" mask (large $\alpha$) and "soft" mask (small $\alpha$). In $M_R$, a value close to 1 indicates static rigid region whereas a value near 0 indicates the presence of a moving area.

% => Figure: Proposed segmentation loss.

$\textbf{Backward}$. RfM can be trained in self-supervision without any ground truth. Since the rigid area is given by $M_R$, the difference between image $I_t$ and its background reconstruction (warped by rigid flow) at rigid area should be zero only when there is no moving object detected in $M_R$. Hence, we could optimize RfM in an end-to-end fashion via minimizing the rigid photometric loss $\mathcal{L}_r$ between $I_t$ and $w_f(I_{t+1}, -F^r_{t\ra{t+1}})$ on $M_R$, in which the Warping function $w_f(I, F)$ bilinearly interpolates image $I$ according to the flow $F$. Formulation details are given in Section \ref{sec:3.4}.

However, simply optimizing $\mathcal{L}_r$ is prone to trivial solutions, where $\mathcal{L}_r \ra 0$ could be satisfied by generating an all(near)-zero rigid map from RfM as illustrated in Fig. \ref{fig:BndLoss} (b). The reason is that RfM tries to minimize $\mathcal{L}_r$ by reducing the number of inliers (rigid pixels). To prevent this, we design a new boundary loss $\mathcal{L}_{bnd}$ to encourage RfM to find out more inliers as much as possible by restricting the area ratio between rigid ($M_R$) and non-rigid ($1-M_R$) regions.
\begin{equation}
    \label{eq:loss:L_bnd}
    \mathcal{L}_{bnd} = \frac{||1-M_R||_1}{||M_R||_1}.
\end{equation}
In (\ref{eq:loss:L_bnd}), $l_1$-norm is designed to approximate the area for soft mask and enable end-to-end differentiability.

% === Section 3.3 === %
\subsection{Overall Structure}
\label{sec:3.3}

% => Figure: overall framework.
\begin{figure*}[htbp]
\begin{center}
    \includegraphics[width=0.85\linewidth]{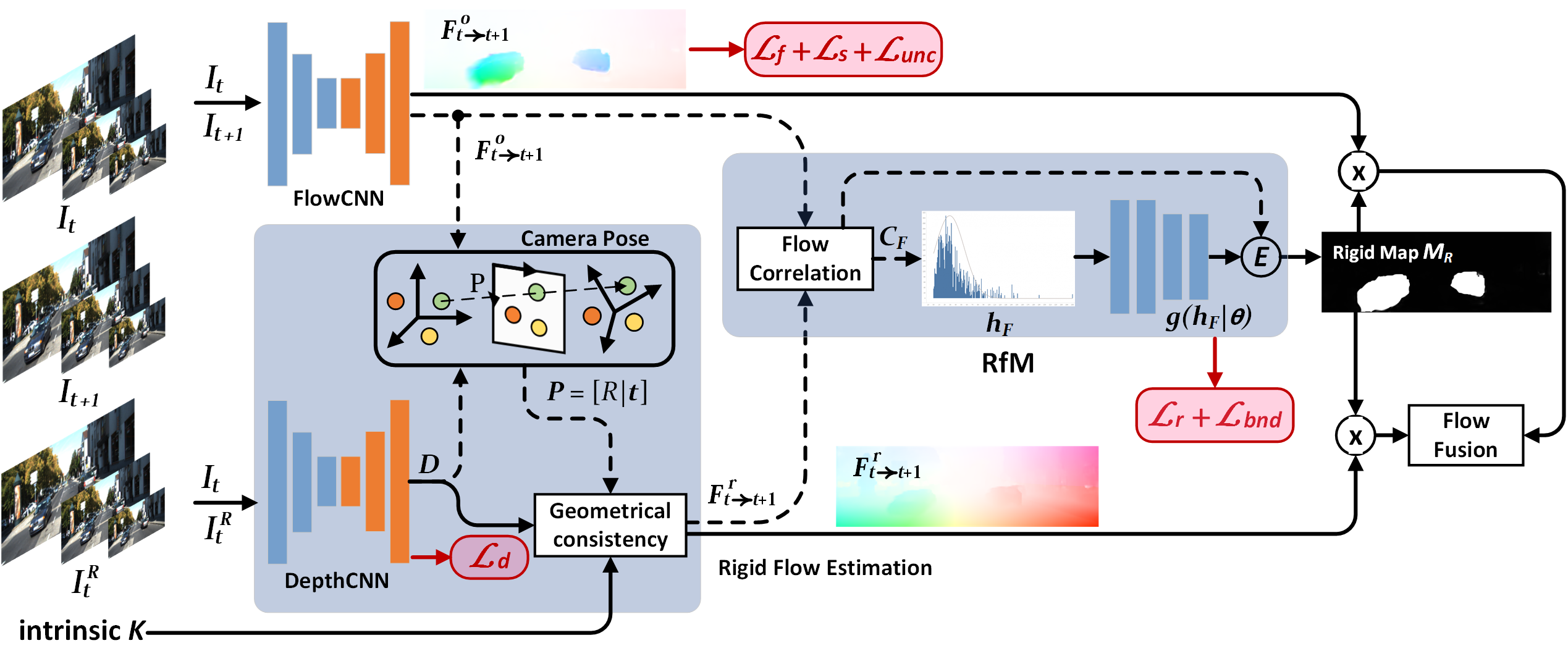}
\end{center}
\vspace{-3mm}
   \caption{Overall architecture of EffiScene. The solid line and dashed line both indicate the forward propagation, but the gradient can only flow back through the solid line during training process due to the non-differentiable operations. Loss functions are shown in the red boxes. }
\label{fig:structure}
\vspace{-5mm}
\end{figure*}

Following the proposed pipeline, we construct our overall EffiScene structure in Fig. \ref{fig:structure}. In our framework, optical flow and depth are estimated by FlowCNN and DepthCNN, respectively. FlowCNN takes two consecutive frames $I_t$ and $I_{t+1}$ as inputs and computes a double channel optical flow $F^o_{t\ra{t+1}}$ representing the horizontal and vertical pixel movements from time $t$ to $t+1$. DepthCNN uses stereo left-right view image pair $I_{t}$ and $I^R_{t}$ at time $t$, and generates a single channel depth map $D$. Any existing deep models can be employed here for FlowCNN and DepthCNN. Camera pose $P=[R|\bm{t}]$ is made up of a rotation matrix $R \in \mathbb{R}^{3\times3}$ and a translation vector $\bm{t} \in \mathbb{R}^{3\times1}$ with respect to the world. Since $F^o_{t\ra{t+1}}$ and $D$ are obtained, $P$ can be computed in (\ref{eq:pnp}) by minimizing the reprojection error derived from (\ref{eq:CameraModel}) between the transformed coordinates $\mathbf{x}_{t+1}=\mathbf{x}_{t}+F^o_{t\ra{t+1}}$ and the projected 3D points $KP\mathbf{X}_{t}=KP[\mathbf{x}_{t}+D]$:
\begin{equation}
    \mathop{\arg\min}_{P} \sum ||[\mathbf{x}_{t+1};1]-\frac{1}{d} KP[\mathbf{X}_{t};1]||_2.
\label{eq:pnp}
\end{equation}
We follow \cite{UnRigidFlow} to solve the $\mathop{\arg\min}$ problem by adopting the Perspective-$n$-Points (P$n$P) method from Simultaneous Localization And Mapping (SLAM) community with a Random Sample Consensus (RANSAC) scheme based on the Levenberg-Marquardt optimization. Once $D$ and $P$ determine the rigid flow $F^r_{t\ra{t+1}}$ via geometrical consistency from (\ref{eq:RigidFlow}), RfM will then adaptively recognize the pixel-wise scene rigidity by jointly considering the two flows. Operator $E$ in Fig. \ref{fig:structure} stands for the outlier exclusion step in RfM. Finally, we refine the fused flow via $F_{t\ra{t+1}}=M_RF^r_{t\ra{t+1}}+(1-M_R)F^o_{t\ra{t+1}}$ for more accurate estimation.

In EffiScene, different modules are closely coupled by RfM, and all components can be clearly interpreted from the geometrical view. Therefore, the efficient rigidity inference can be carried out via jointly considering the geometrical information from flow and depth.

% === Section 3.4 === %
\subsection{Losses and Regularization}
\label{sec:3.4}
Photometric error evaluates the photometric similarity between two images $I$ and $\hat{I}$ as defined in (\ref{eq:loss:photometric}), in which $\lambda_\rho$ balances the $l_1$-norm and SSIM term \cite{SSIM}:
\begin{equation}
\setlength{\abovedisplayskip}{3pt}
\setlength{\belowdisplayskip}{3pt}
    \rho(I, \hat{I}) = \lambda_\rho l_1(I-\hat{I}) + (1-\lambda_\rho) {\rm {SSIM}} (I, \hat{I}).
\label{eq:loss:photometric}
\end{equation}
We design different loss functions to train EffiScene in an unsupervised manner based on photometric error.

\noindent
$\textbf{Optical Flow Loss}$. Optical flow from FlowCNN is optimized by minimizing the photometric error between the original image $I_t$ and its reconstruction $\hat{I^o_t}$ from optical flow $F^o_{t\ra{t+1}}$ on non-occluded region $M_{noc}$, which is determined by forward-backward flow check \cite{DF-Net}.
\begin{equation}
\setlength{\abovedisplayskip}{3pt}
\setlength{\belowdisplayskip}{3pt}
   \mathcal{L}_f = \frac{1}{\sum {M_{noc}}}\sum\nolimits_{\Omega} M_{noc} \cdot \rho (I_t, \hat{I^o_t}).
\label{eq:loss:L_f}
\end{equation}
Additionally, edge-aware smooth loss is used to regularize the optical flow on the full image domain $\Omega$.
\begin{equation}
\setlength{\abovedisplayskip}{3pt}
\setlength{\belowdisplayskip}{3pt}
    \mathcal{L}_s = \sum\nolimits_{\Omega} |\triangledown^2F^o_{t \ra {t+1}}| e^{-|\triangledown^2I_t|}.
\label{eq:loss:L_s}
\end{equation}
We use $2^{nd}$ order gradient to eliminate the velocity impact.

\noindent
$\textbf{Depth Loss}$. Similar to optical flow, depth map from DepthCNN is trained with photometric loss and smooth loss, but for stereo pairs instead, e.g. left-view image $I_t$ and synthesized image $\tilde{I_t}$ from right-view frame $I^R$. Left-right consistency from Godard \etal \cite{MonoDepth} is also adopted as penalty to ensure the stereo depth coherence below
\begin{equation}
\setlength{\abovedisplayskip}{3pt}
\setlength{\belowdisplayskip}{3pt}
    \mathcal{L}_d = \sum\nolimits_{\Omega} \rho (I_t, \tilde{I_t}) + |\triangledown^2D| e^{-|\triangledown^2I_t|} + |D - \tilde{D}^L|,
\label{eq:loss:L_d}
\end{equation}
where $\tilde{D}^L$ is the projected left-view depth from the right.

\noindent
$\textbf{RfM Loss}$. As discussed in Sec. \ref{sec:3.2}, the boundary loss $\mathcal{L}_{bnd}$ in (\ref{eq:loss:L_bnd}) and the rigid photometric loss $\mathcal{L}_r$ in (\ref{eq:loss:L_r}) are both used to train RfM. In $\mathcal{L}_r$, error between $I_t$ and rigid reconstruction $\hat{I^r_t}$ from rigid flow $F^r_{t\ra{t+1}}$ are evaluated on the rigid map $M_R$ as
\begin{equation}
\setlength{\abovedisplayskip}{3pt}
\setlength{\belowdisplayskip}{3pt}
    \mathcal{L}_r = \frac{1}{\sum M_R} \sum\nolimits_{\Omega} M_R \cdot \rho (I_t, \hat{I^r_t}).
\label{eq:loss:L_r}
\end{equation}

\begin{figure}[h]
\begin{center}
    \includegraphics[width=0.95\linewidth]{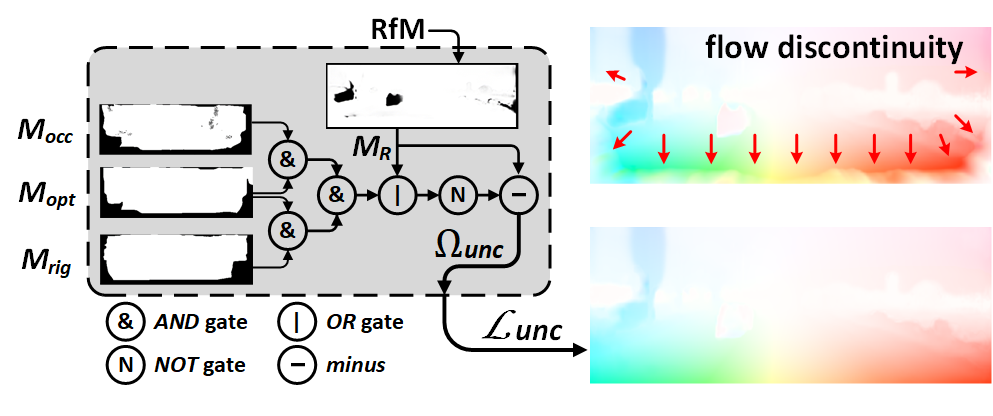}
\end{center}
    \caption{Regularizing optical flow for boundary discontinuity.}
\label{fig:unc_loss}
\end{figure}

\noindent
$\textbf{Regularization.}$ Minimizing $\mathcal{L}_f$ may cause the discontinuity of optical flow at image boundary due to the uncovered area between two frames. We use rigid flow to rectify the optical flow by enforcing flow consistency based on the learned rigid map $M_R$ from RfM. However, $M_R$ might cover undesired moving objects at the beginning of the training. We improve $M_R$ via generating robust uncover region $\Omega_{unc}$ by fusing occlusion mask ($M_{occ}$), valid optical flow mask ($M_{opt}$) and valid rigid flow mask ($M_{rig}$) as shown in Fig. \ref{fig:unc_loss}. 
% => Figure: UnCover loss results demo.
The non-occluded region is defined as $M_{occ}$ while $M_{opt}$ and $M_{rig}$ indicate valid motion from optical flow and rigid flow.
Next, the uncover loss for flow regularization is defined as:
\begin{equation}
    \mathcal{L}_{unc} = \sum\nolimits_{\Omega_{unc}} {||F^o_{t \ra {t+1}}-F^r_{t \ra {t+1}}||}^2_2.
\label{eq:loss:uncoverloss}
\end{equation}

All losses are combined to train EffiScene as an energy minimization optimization as follows:
\begin{small}
\begin{equation}
    E = \lambda_{f} \mathcal{L}_{f} + \lambda_{s} \mathcal{L}_{s} + \lambda_{r} \mathcal{L}_r + \lambda_{d} \mathcal{L}_d + \lambda_{bnd} \mathcal{L}_{bnd} +  \lambda_{unc} \mathcal{L}_{unc},
\label{eq:loss:all}
\end{equation}
\end{small}
where $\lambda_{f/s/r/d/bnd/unc}$ provide the weighting trade-offs.

%%%%%%%%% ==> Experiments
\section{Experiments}
\label{sec:4}
Extensive experiments are conducted to benchmark EffiScene against SOTA scene flow methods in four sub-tasks: \emph{(i)} optical flow estimation; \emph{(ii)} depth prediction; \emph{(iii)} visual odometry; and \emph{(iv)} motion segmentation. Qualitative results are illustrated in Fig. \ref{fig:MainResults}. More results are listed in the Supplementary Materials.

% => Table: qualitative results
\begin{figure*}[htbp]
\begin{center}
    \includegraphics[width=0.95\linewidth]{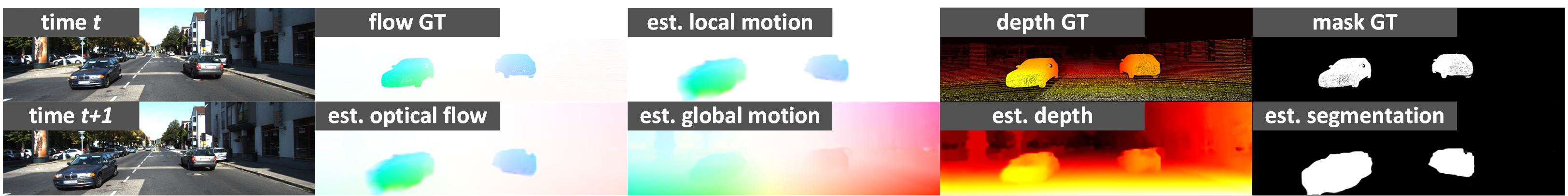}
\end{center}
   \caption{Qualitative results of the proposed method for scene flow estimation (denoted by 'est.'.).}
\label{fig:MainResults}
\end{figure*}

% ===> Implementation Details
\subsection{Implementation Details}
\label{sec:4.1}

\textbf{Dataset.} To keep consistency with previous works \cite{GeoNet, DF-Net, CC, UnOS, UnRigidFlow, MonoDepth, EPC, EPC++}, we use the same dataset and protocol for all experiments. Specifically, 28,968 images out of (42,382) images in KITTI raw set \cite{KITTI_raw} are used to train EffiScene, except for the scenes enrolled in KITTI 2015 \cite{KITTI_2015} training set, which is reserved for optical flow validation as well as depth estimation and motion segmentation with corresponding ground truth. Besides KITTI 2015, KITTI 2012 \cite{KITTI_2012} is also adopted for optical flow evaluation. Different from KITTI 2015, dynamic scenes in KITTI 2012 only contains camera movements but no moving cars. For visual odometry task, we fine-tune our model on sequences 00-08 in KITTI Odometry split \cite{KITTI_2012}, then test it on sequences 09 and 10.

\textbf{Network deployment.} For FlowCNN, we employ RAFT \cite{RAFT} as the baseline due to its excellent performance in supervised optical flow estimation and make a few modifications for the unsupervised setting. We also modify PWC-Net \cite{PWC-Net} for DepthCNN by changing the output of the last convolutional layer from 2 channels to 1 channel to generate a single channel depth map, and replace the \emph{deconv} layer by bilinear upsampling to avoid the checkerboard artifacts. In RfM, $g(f_F|\theta)$ is obtained from two fully connected layers with size 100-32 and 32-1 followed by ReLU and Sigmoid activation, respectively.

\textbf{Training.} We train EffiScene from scratch in three stages without any ground truth. By default, photometric balance $\lambda_{\rho}$ in (\ref{eq:loss:photometric}) is set to 0.003 for all experiments. Weighting for loss functions denoted by $\{\lambda_{f}, \lambda_{s}, \lambda_{r}, \lambda_{d}, \lambda_{bnd}, \lambda_{unc} \}$ in (\ref{eq:loss:all}) are initialized to all zeros, then we adjust them in different stages. In the first stage, we train FlowCNN and fix DepthCNN and RfM to obtain a coarse optical flow $F^o_{t \ra {t+1}}$ by setting $\lambda_{f}=1.0$ and $\lambda_{s}=0.5$ for 20 epochs. In parallel, we train DepthCNN independently for depth $D$ by setting $\lambda_{d}=1.0$ for 50 epochs as suggested in \cite{MonoDepth}. Once we achieve a reasonable optical flow and depth, we fix FlowCNN and DepthCNN, and train RfM for 10 epochs in the second stage. Here, we set $\lambda_{r}=1.0$ and $\lambda_{bnd}=0.023$, and others to zeros. Finally, in the last stage, we jointly fine-tune all the networks based on the fused flow $F_{t \ra {t+1}}$ by setting $\{\lambda_{r}, \lambda_{d}, \lambda_{bnd}, \lambda_{unc} \}=\{1.0, 1.0, 0.023, 1.0 \}$ for 10 epochs.

All input images are resized to $256 \times 832$, and the AdamW optimizer \cite{AdamW} is utilized for optimization with momentum [0.9, 0.99] and weight decay 1e-5. Batch size is set to 4. The initial learning rate is first set to 1e-4 for the first two training stages, reduced to 1.25e-5 for the last stage, and it is decreased by a factor of 2 for every 50K batch. All models are trained on a single Tesla P40 GPU for about 150 GPU hours. Unlike existing methods \cite{UnRigidFlow, UnOS} whose performances are highly relied on the prefixed thresholds, there is no empirical parameter needed to be set by the user in EffiScene in both training and testing.

% ===> Evaluation
\subsection{Evaluation}
\label{sec:4.2}

% ===> Table for optical flow comparisons
\begin{table*}[htbp]
\small
\begin{spacing}{1.0}
\begin{center}
    \caption{Quantitative results of optical flow estimation. Averaged end-point-error (EPE) is used for evaluation except for the last two columns which tabulate the percentage of erroneous pixels (Fl-all). 'Noc' and 'Occ' mean non-occluded region and occluded region.}
\label{tab:FlowEval}
\setlength{\tabcolsep}{1.45mm}{
\begin{tabular}{p{2.5cm}<{\centering} p{0.9cm}<{\centering} p{0.9cm}<{\centering} | p{1cm}<{\centering} p{1cm}<{\centering} p{1cm}<{\centering} | p{1cm}<{\centering} p{1cm}<{\centering} p{1cm}<{\centering} p{1cm}<{\centering} p{1cm}<{\centering}}
    \bottomrule[2pt]
    \multirow{3}{1.0cm}{\centering Method} & & & \multicolumn{3}{c|}{KITTI 2012} & \multicolumn{5}{c}{KITTI 2015} \\
    & Stereo & \multirow{2}{0.9cm}{\centering Super- vised} & \multicolumn{3}{c|}{Train Average EPE} & \multicolumn{3}{c}{Train Average EPE} & \multirow{2}{1cm}{\centering Train Fl-all} & \multirow{2}{1cm}{\centering Test Fl-all}\\
    \cline{4-9}
    & & & Noc & Occ & All & Move & Static & All \\
    \hline
    FlowNet2 \cite{FlowNet2}      & & \checkmark & - & - & 4.09 & - & - & 10.06 & 30.37\% & - \\
    PWC-Net \cite{PWC-Net}        & & \checkmark & - & - & 4.14 & - & - & 10.35 & 33.67\% & - \\

    UnFlow-CSS \cite{UnFlow}      & & & 1.26 & - & 3.29 & - & - & 8.10 & 23.27\% & - \\
    DF-Net \cite{DF-Net}          & & & - & - & 3.54 & - & - & 8.98 & 26.01\% & 25.70\% \\
    Self-Mono-SF \cite{SelfMonoSF}& & & - & - & - & - & - & 7.51 & 23.49\% & 23.54\% \\
    CC \cite{CC}                  & & & - & - & - & 5.67 & 5.04 & 6.21 & 26.41\% & - \\
    CC-uft \cite{CC}              & & & - & - & - & - & - & 5.66 & 20.93\% & 25.27\% \\
    EPC++ \cite{EPC++}            & \checkmark & & - & - & 1.91 & - & - & 5.43 & - & 20.52\% \\
    UnOS \cite{UnOS}              & \checkmark & & \textbf{1.04} & 5.18 & \textbf{1.64} & 5.30 & 5.39 & 5.58 & - & 18.00\% \\
    UnRigidFlow \cite{UnRigidFlow} & \checkmark & & 1.09 & 4.87 & 1.92 & 7.92 & 3.85 & 5.19 & 14.68\% & \textbf{11.66\%} \\

    \hline
    EffiScene (-pwc) & \checkmark & & 1.19 & 4.74 & 1.71 & 7.63 & 3.72 & 4.92 & 14.55\% & - \\
    EffiScene        & \checkmark & & 1.19 & \textbf{4.71} & 1.68 & \textbf{5.15} & \textbf{3.69} & \textbf{4.20} & \textbf{14.31\%} & 13.08\% \\

\toprule[1.5pt]
\end{tabular}}
\end{center}
\end{spacing}
\vspace{-5mm}
\end{table*}

% ===> Evaluate: Optical flow
\textbf{Optical flow estimation.} Optical flow comparisons with supervised and unsupervised methods are summarized in Tab. \ref{tab:FlowEval}. On KITTI 2015, our method achieves the best performances on averaged end-point-error (EPE) across all image regions, e.g. moving area, static area $\ldots$ Specifically, for the most vital metric EPE-All, EffiScene significantly reduces the existing error by a considerable margin from 5.19 \cite{UnRigidFlow} to 4.20 ($19.1\%$ relative improvement). We also achieve the best and the second best Fl-all error $14.31\%$ and $13.08\%$ among all approaches on training and testing set, respectively. On KITTI 2012 dataset, EffiScene consistently surpasses UnRigidFlow \cite{UnRigidFlow} by $12.5\%$ relative EPE growth (1.68 vs. 1.92), which validates the generalization ability of our method. Unfortunately, since there is no moving object in KITTI 2012, the learned rigidity mask $M_R$ from RfM will cover almost the full image, leading to that the fused flow $F_{t \ra {t+1}}$ will be dominated by the rigid flow $F^r_{t \ra {t+1}}$ (from depth and pose). Hence, it is challenging for EffiScene to benefit from FlowCNN, resulting in a slight drop of EPE to 1.68, comparing to the best achievable EPE=1.64 \cite{UnOS}. However, for occluded region, motion can be better inferred by more accurate depth and pose, and we obtain the best result EPE-Occ of 4.71 (vs. 5.18 \cite{UnOS}).  In addition, we also design a variation model EffiScene (-pwc) based on the popular PWC-Net \cite{PWC-Net} backbone for further evaluation. PWC-based EffiScene also achieves SOTA results for both datasets, demonstrating the consistency as well as robustness of the proposed framework.

\textbf{Ablation.} Optical flow from different training stages are listed in Tab. \ref{tab:Ablation}. It is not surprising that optical flow $F^o$ from the 1st training stage yields the worst performance since the geometrical rigidity constraint has not been considered. With the help of RfM, in the 3rd stage, FlowCNN and DepthCNN are jointly optimized based on specific rigid regions $M_R$, and lower errors can be achieved in moving area (EPE-Move=3.09) as well as static regions (EPE-Static=2.09). By jointly fusing $F^o$ and $F^r$, the final output $F$ generates a much better result for all regions with EPE=4.20 and Fl-all=14.31\%.

\begin{table}[htbp]
\small
\begin{spacing}{1.0}
\begin{center}
    \caption{Ablation study on optical flow. Subscript $_{t \ra {t+1}}$ of $F^o_{t\ra{t+1}}$, $F^r_{t\ra{t+1}}$ and $F_{t\ra{t+1}}$ has been omitted for clarity.}
\label{tab:Ablation}
\setlength{\tabcolsep}{1.45mm}{
\begin{tabular}{c p{0.8cm}<{\centering} p{0.8cm}<{\centering} p{0.8cm}<{\centering} p{0.8cm}<{\centering} p{1cm}<{\centering}}
    \bottomrule[1.5pt]
    \multirow{2}{1.5cm}{\centering Flow Type} & \multirow{2}{0.6cm}{\centering Train Stage} & \multicolumn{3}{c}{Train Average EPE} & \multirow{2}{1cm}{\centering Train Fl-all} \\
    \cline{3-5}
    & & Move & Static & All \\

    \hline

    $F^o$ (FlowCNN)    & 1st & 4.38 & 6.80   & 6.76 & 18.89\% \\
    $F^o$ (FlowCNN)    & 3rd & \textbf{3.09} & 4.81 & 4.70 & 15.87\% \\
    $F^r$ (DepthCNN)   & 3rd & 38.10         & \textbf{2.09} & 10.42 & 21.44\% \\
    $F$ (EffiScene)    & 3rd & 5.15          & 3.69 & \textbf{4.20} & \textbf{14.31\%} \\

    \toprule[1.5pt]
\end{tabular}}
\end{center}
\end{spacing}
\vspace{-5mm}
\end{table}

% ===> Evaluate: Depth
\textbf{Stereo-depth prediction.} Depth estimation is evaluated on KITTI train set with standard metrics \cite{Vid2Depth} in Tab. \ref{tab:DepthEval}. By jointly considering the inherent relation between flow, pose, rigidity and depth, superior depth maps can be predicted comparing with both monocular or stereo based methods. Surprisingly, EffiScene even outperforms SsSMnet \cite{SsSMnet} and MonoDepth \cite{MonoDepth} which are specifically designed for the depth estimation task. Since there is no new depth-specific components designed in EffiScene, we hypothesize that the gain in depth estimation may come from the collaborative training process, where optical flow, camera pose and depth are coupled by RfM for mutual reinforcement.

% ===> Table for depth comparisons
\begin{table*}[htbp]
\small
\begin{spacing}{1.0}
\begin{center}
    \caption{Quantitative results of depth estimation conducted on the KITTI 2015 training set. Depth errors (middle columns) and prediction accuracy (right columns) are used for evaluation. All valid depth ranges are capped at 80m.}
\label{tab:DepthEval}
\setlength{\tabcolsep}{1.45mm}{
\begin{tabular}{p{3.0cm}<{\centering} p{1.2cm}<{\centering} |  p{1.2cm}<{\centering} p{1.2cm}<{\centering} p{1.2cm}<{\centering} p{1.2cm}<{\centering} | p{1.2cm}<{\centering} p{1.2cm}<{\centering} p{1.2cm}<{\centering}}

    \bottomrule[1.5pt]
    \multirow{2}{1cm}{\centering Method} & \multirow{2}{1cm}{\centering Stereo} & \multicolumn{4}{c|}{Error (lower is better)} & \multicolumn{3}{c}{Accuracy, $\delta$ (higher is better)}\\
    & & AbsRel & SqRel & RMSE & RMSlog & $< 1.25$ & $< {1.25}^2$ & $< {1.25}^3$ \\

    \hline
    % sfMLearner \cite{sfMLearner} & & 0.208 & 1.768 & 6.856 & 0.283 & 0.678 & 0.885 & 0.957\\
    % DF-Net \cite{DF-Net}         & & 0.150 & 1.124 & 5.507 & 0.223 & 0.800 & 0.933 & 0.973\\
    CC \cite{CC}                   & & 0.140 & 1.070 & 5.326 & 0.217 & 0.826 & 0.941 & 0.975\\
    Self-Mono-SF \cite{SelfMonoSF} & & 0.125 & 0.978 & 4.877 & 0.208 & 0.851 & 0.950 & 0.978\\

    % MonoDepth \cite{MonoDepth} & \checkmark& 0.124 & 1.388 & 6.125 & 0.217 & 0.841 & 0.936 & 0.975\\
    EPC \cite{EPC}             & \checkmark & 0.109 & 1.004 & 6.232 & 0.203 & 0.853 & 0.937 & 0.975\\
    EPC++ \cite{EPC++}         & \checkmark & 0.127 & 0.936 & 5.008 & 0.209 & 0.841 & 0.946 & 0.979\\
    % UnRigidFlow \cite{UnRigidFlow} & \checkmark & 0.108 & 1.020 & 5.528 & 0.195 & 0.863 & 0.948 & 0.980\\

    SsSMnet \cite{SsSMnet}     & \checkmark & 0.075 & 1.726 & 4.857 & 0.165 & 0.956 & 0.976 & 0.985\\
    MonoDepth \cite{MonoDepth} & \checkmark & 0.068 & 0.835 & 4.392 & 0.146 & 0.942 & 0.978 & 0.989\\
    UnRigidFlow \cite{UnRigidFlow} & \checkmark & 0.051 & 0.532 & 3.780 & 0.126 & 0.957 & 0.982 & 0.991\\

    EffiScene & \checkmark & \textbf{0.049} & \textbf{0.522} & \textbf{3.461} & \textbf{0.120} & \textbf{0.961} & \textbf{0.984} & \textbf{0.992}\\
    \toprule[1.5pt]

\end{tabular}}
\end{center}
\end{spacing}
\vspace{-6mm}
\end{table*}

% ===> Evaluate: Motion segmentation.
\textbf{Motion segmentation.} Results from RfM is also used to evaluate motion segmentation with advanced scene flow approaches as listed in Tab. \ref{tab:SegmentEval}. We achieve the best pixel accuracy and mean accuracy by surpassing the baseline UnOS \cite{UnOS} 4.5\% and 2.8\%. However, considering that moving cars just make up only a small portion of the full image in KITTI 2015 (usually less than 5\%), high accuracy does not always imply a superior segmentation ability due to severe class imbalance. Therefore, Intersection-Over-Union (IoU) could be a fairer and more compelling benchmark as listed in the last two columns of Tab. \ref{tab:SegmentEval}. Our method improves the mean IoU from SOTA 0.570 \cite{UnRigidFlow} to 0.615, and the frequency weighted (f.w.) IoU from 0.900 to 0.926. Note that
CC \cite{CC} adopts a much more complex and deep auto-encoder for segmentation, but it is still 4.6\% lower than the proposed method because of the independent segmentation inference structure.

% ===> Table for motion segmentation comparisons
\begin{table}[htbp]
\small
\begin{spacing}{1.0}
\begin{center}
    \caption{Quantitative results of motion segmentation. IoU based metrics (last two columns) are more meaningful for KITTI 2015.}
\label{tab:SegmentEval}
\setlength{\tabcolsep}{1.45mm}{
\begin{tabular}{c p{1cm}<{\centering} p{1cm}<{\centering} p{1cm}<{\centering} p{1cm}<{\centering}}
    \bottomrule[1.5pt]
    \multirow{2}{1cm}{\centering Method} & Pixel Acc. & Mean Acc. & Mean IoU & f.w. IoU \\

    \hline
    EPC \cite{EPC}                 & 0.890 & 0.750 & 0.520 & 0.870 \\
    EPC++ \cite{EPC++}             & 0.910 & 0.760 & 0.530 & 0.870 \\
    UnOS (full) \cite{UnOS}        & 0.900 & 0.820 & 0.560 & 0.880 \\
    CC \cite{CC}                   &   -   &   -   & 0.569 &   -   \\
    UnRigidFlow \cite{UnRigidFlow} & 0.930 & 0.840 & 0.570 & 0.900 \\

    EffiScene & \textbf{0.945} & \textbf{0.848} & \textbf{0.615} & \textbf{0.926} \\

    \toprule[1.5pt]
\end{tabular}}
\end{center}
\end{spacing}
\vspace{-5mm}
\end{table}

% ===> Evaluate for camera pose
\textbf{Visual odometry.} Absolute Trajectory Error (ATE) in \cite{ORB-SLAM, sfMLearner} is utilized for camera pose evaluation in Tab. \ref{tab:PoseEval}. Two types of technical strategies are summarized: Deep Neural Netwrok (DNN) based and P$n$P based pose estimation. Usually, DNN based methods run faster than optimization-based P$n$P, but offer lower accuracy \cite{DDVO, Reza, UnOS}. Since only 2 frames are used in our method, we follow \cite{UnOS, UnRigidFlow} and average the accumulated poses from neighboring 5 frames for fair comparison with multi-frame competitors. For P$n$P based methods, we outperform all existing methods in both sequences. Note that the performance of P$n$P heavily depends on the quality of predicted flow and depth, better optical flow and depth estimation from previous steps will definitely contribute to improvement in camera pose accuracy. For DNN approaches, we also give a variation model EffiScene (-dnn) for fair comparison -- replacing the P$n$P method with a 9-layer fully convolutional network with channels [16, 32, 64, 128, 256*4, 6] to regress the 6-DoF pose matrix $P=[R|\bm{t}]$. It achieves competitive results by surpassing \cite{sfMLearner} and \cite{DF-Net}, but it narrowly trails \cite{GeoNet} and \cite{CC}, both having access to more frames than us.

% ====> Table for pose estimation comparisons
\begin{table}[htbp]
\small
\begin{spacing}{1.0}
\begin{center}
    \caption{Comparisons of visual odometry on KITTI Odometry.}
\label{tab:PoseEval}
\setlength{\tabcolsep}{1mm}{
\begin{tabular}{p{2.4cm}<{\centering} p{0.8cm}<{\centering} p{0.8cm}<{\centering} p{1.65cm}<{\centering} p{1.65cm}<{\centering}}
    \bottomrule[1.5pt]
    \multirow{2}{*}{\centering Method} & \multirow{2}{0.8cm}{\centering Frames} & \multirow{2}{0.8cm}{\centering Pose} & \multirow{2}{1cm}{\centering Sequence 09} & \multirow{2}{1cm}{\centering Sequence 10}\\
    \\

    \hline
    DF-Net \cite{DF-Net}           & 5 & DNN & 0.017$\pm$0.007 & 0.015$\pm$0.009 \\
    sfMLearner \cite{sfMLearner}   & 5 & DNN & 0.016$\pm$0.009 & 0.013$\pm$0.009 \\
    GeoNet \cite{GeoNet}           & 5 & DNN & 0.012$\pm$0.007 & 0.012$\pm$0.009 \\
    CC \cite{CC}                   & 5 & DNN & \textbf{0.012}$\pm$\textbf{0.007} & \textbf{0.012}$\pm$\textbf{0.008} \\
    EffiScene (-dnn)               & 2 & DNN & 0.013$\pm$0.006 & 0.013$\pm$0.008 \\

    \hline

    ORB-SLAM \cite{ORB-SLAM}       & all & P$n$P & 0.014$\pm$0.008 & 0.012$\pm$0.011 \\
    Vid2Depth \cite{Vid2Depth}     & 3 & P$n$P & 0.013$\pm$0.010 & 0.012$\pm$0.011 \\
    UnOS \cite{UnOS}               & 2 & P$n$P & 0.012$\pm$0.006 & 0.013$\pm$0.008 \\
    UnRigidFlow \cite{UnRigidFlow} & 2 & P$n$P & 0.012$\pm$0.007 & 0.012$\pm$0.006 \\
    EffiScene                      & 2 & P$n$P & \textbf{0.011}$\pm$\textbf{0.006} & \textbf{0.011}$\pm$\textbf{0.008} \\

    \toprule[1.5pt]
\end{tabular}}
\end{center}
\end{spacing}
\vspace{-5mm}
\end{table}

% ===> Figure for error regions from RfM
\begin{figure}[htbp]
\begin{center}
\includegraphics[width=0.80\linewidth]{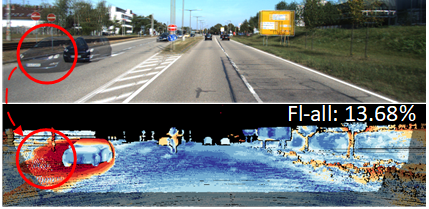}
\end{center}
    \caption{High errors caused by inaccurate $R_M$ and occlusion.}
\label{fig:ErrRegions}
\vspace{-5mm}
\end{figure}

%% ===> Complexity & Limitation Analysis
\subsection{Analysis}
\textbf{Complexity analysis.} Running time and model size are listed in Tab. \ref{tab:TimeEval}. Our method runs more than 2 times faster than UnOS, but is slower than CC due to the time consuming P$n$P step as discussed in Sec. \ref{sec:4.2}. By replacing P$n$P with a deep network, EffiScene (-dnn) significantly speeds up the inference with acceptable performance drop (0.002 drop from Tab. \ref{tab:PoseEval}). EffiScene requires much fewer number of learnable parameters in the full model (\#Params) and the segmentation module (\#SegParams).

% ===> Table for running time
\begin{table}[htbp]
\small
\begin{spacing}{1.0}
\begin{center}
    \caption{Model complexity analysis. Experiments are performed on the same computing platform with a single Tesla P40 GPU.}
\label{tab:TimeEval}
\setlength{\tabcolsep}{1.0mm}{
\begin{tabular}{p{2.4cm}<{\centering} p{1.15cm}<{\centering} p{0.75cm}<{\centering} p{1.25cm}<{\centering} p{1.75cm}<{\centering}}
    \bottomrule[1.5pt]
    \multirow{2}{*}{\centering Method} & RunTime (ms) & FPS (f/s) & \#Params (Mb) & \#SegParams (Mb) \\

    \hline

    CC \cite{CC}                   & 49.55  & 20.18 & 74.26 & 5.22 \\
    UnOS \cite{UnOS}               & 228.16 & 4.38 & 17.06 & - \\
    UnRigidFlow \cite{UnRigidFlow} & 87.57  & 11.42 & \textbf{10.22} & - \\
    EffiScene & 93.11 & 10.74 & 10.36 & \textbf{0.0032} \\
    EffiScene (-dnn) & \textbf{47.06} & \textbf{21.25} & 12.54 & \textbf{0.0032} \\

    \toprule[1.5pt]
\end{tabular}}
\end{center}
\end{spacing}
\vspace{-5mm}
\end{table}

% ===> Limitations
\textbf{Limitations. }
Although promising EPE (=4.20) is achieved in our model, test Fl-all error (13.08\%) is still 1.42\% higher than \cite{UnRigidFlow} as depicted in Tab. \ref{tab:FlowEval}. This indicates that EffiScene learns better optical flow for those 'good' regions (lower EPE), but not for all pixels (higher Fl). One reason is that the learned rigidity map $M_R$ could be wrongly estimated at occluded regions, where local motion is difficult to optimize due to missing pixels, resulting in unreliable motion correlation for RfM. For example, in Fig. \ref{fig:ErrRegions}, backgrounds occluded by the moving cars have much higher errors. Therefore, improving RfM in the large occlusion case seems to be a logical next step.

%%%%%%%%% ==> Conclusion
\section{Conclusion}
In summary, we propose EffiScene for unsupervised scene flow estimation by coupling several low-level vision sub-tasks. We demonstrate that per-pixel rigidity can be efficiently inferred by jointly exploiting optical flow, depth and camera pose, since they share the same inherent geometrical structure with scene rigidity. By exploring joint rigidity learning, more accurate rigid constraint and efficient network training can be achieved. Extensive experiments on scene flow benchmarks produce SOTA results with simpler model for all four sub-tasks, demonstrating the effectiveness of the proposed method.
In our future work, long term dependency will be explored in EffiScene to solve the rigidity inference with large occlusions.

{\small
\bibliographystyle{ieee_fullname}
\bibliography{SceneFlowRefs}
}

\end{document}